# An Event-triggered System for Social Persuasion and Danger Alert in Elder Home Monitoring

*Jun-Yi Liu* (劉俊毅), *Chung-Hao Chen* (陳君皓), *Ya-Chi Tsao* (曹雅淇), *Ssu-Yao Wu* (吳思瑤),

*Yu-Ting Tsao* (曹瑜庭), *and \*Lyn Chao-ling Chen* (陳昭伶)

Department of Information Management,
Fu Jen Catholic University, New Taipei City, Taiwan
\*E-mail: lynchen@ntu.edu.tw

**ABSTRACT**

In the study, the physical state and mental state of elders are both considered, and an event-triggered system has developed to detect events: watch dog, danger notice and photo link. By adopting GMM background modeling, the motion behavior of visitors and elders can be detected in the watch dog event and danger notice event respectively. Experiments set in home scenarios and 5 families participated in the experiments for detecting and recording three types of events from their life activities. In addition, the captured images were analyzed using SVM machine learning. For lack of technical experiences of elders, an intuitive operation as normal life activity was designed to create communication between elder and relatives via social media.

***Keywords:*** *Background modeling, Event-Triggered, Image sequence analysis, Motion detection, Shape recognition, Social media, Social persuasion, Surveillance*

## 1. INTRODUCTION

The health of growing elder group has been focused in the modern society. While young people work outside, they often stay home alone. From the data of Ministry of Health and Welfare, the interaction rate of people gradually decreases in the aging process, and not only safety of elders but also their social behavior is considered in the study.

**1.1 Psychological condition of elders**

The usual problem reflects in psychological condition of elders is few interactions with others. Especially, elders who live alone tend to have someone to accompany them, and they often worry about passing away without anyone knowing. The lack of security and mental support is emotionally distressed, and results to mental illness or even worse situation [1]. What is valuable in the social behavior of elders not the effectiveness or frequency of social communication, but the self-awareness of warmth and satisfaction of personal needs [1]. From the interview, the elders talked about that "Friends and neighbors only have time to chat, and will not talk about any inner words," "We do not take the initiative to disturb them and we're embarrassed to trouble others" [1]. People often have less patience to listen to elders in their families, and ignore their psychological problems [2]. Most of them live alone and do not arrange leisure activities that generates the lonely feel of outside world, and they do not like to interact with people, etc. [2]. The study also pointed that many elders live alone by for the limitation of environment, aging of physical functions also limits their opportunities for social contact, and causes the difficulty to maintain or expand the original social circle [2]. There is significant positive correlation between social support and personality and life satisfaction of elders who live alone [3]. For improving the situations and avoiding depression and relieve stress of elders, many scholars have claimed that appropriate social activities can not only help to improve loneliness and social isolation dilemmas of elders, but also bring them satisfaction from interactions with friends or families. In addition, it also helps to maintain their physiological functions, and increase muscle tone to maintain physical strength and to stimulate degenerate functions. For the reason, social interaction has great influence on the physical and mental health of elders. Hence, psychological condition of elders is considered in the study.

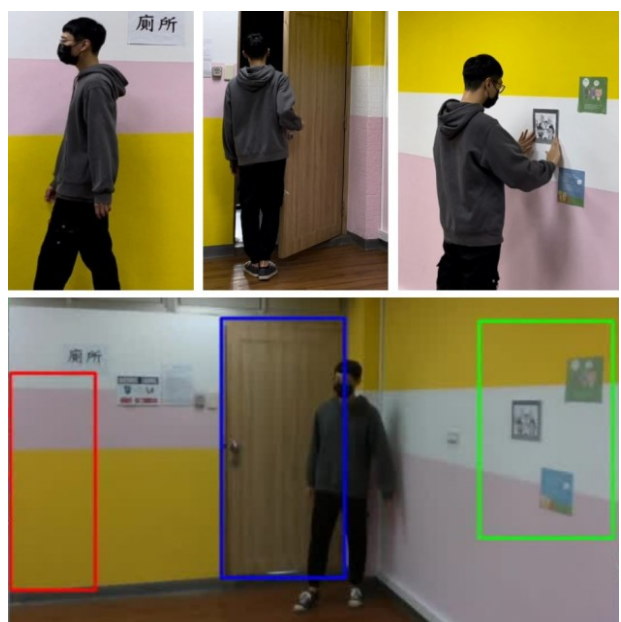

Fig. 1. Three types of events in the monitoring system: watch dog event (blue area), danger notice event (red area) and photo link event (green area), and the correlated scenarios of events on the top from left to right.

## 1.2 Physical condition of elders

Considering the physical condition, elders often have accidents on the stairs or in the bathroom for the deterioration of physical functions. One common danger event of elders is fall accident at home, and often causes serious injury such as a fractured. Family members have to pay attention to their health, for muscle weakness and out of control of their bodies may cause great danger in their daily lives. The conscious health outcomes declared that most elders think their health state remain almost the same or decline [4]. In addition, the economic condition affects the dignity of individuals and the overall quality of life, and shows that elders who live alone with good economic conditions can face life positively [4]. However, most of them have retired with no monetary income in the modern society, and do not have independent ability. While family members work or study outside, elders often stay at home alone with no care in time, and increase the possibilities of accidents, such as fall or hurt [5]. The common places where accidents happen are doorsills, high stairs and a wet bathroom [5]. Actually, elders ask for help initiatively when they fall or have abnormal body index such as blood pressure or blood sugar, and they need timely assistance [2]. If there is no one around them for providing appropriate help in time, and that may aggravate the situation [2]. Hence, physical condition of elders is considered in the study.

## 2. RELATED WORKS

### 2.1. Social promotion of elders

Social interaction helps to keep mental health of elders, however, only few researches focused on the issue and it brings advantage in the study. The system has developed to promote social interaction for improving mental health of elders. Social persuasion systems for elders were classified into two categories: interactive application and physical device. Photo Alive is a mobile application to provide social interaction function by sharing photos, and the photo collection connects elders and their friends or family members [6]. For most elders are not familiar with complex operation of mobile application, Berbakti designed a simple interface for elders sending messages to their relative, and an additional function used alerts to remind them have not replay messages in a period [7]. Another application has been implemented in the SAFER platform, and record personal information and favorite activities of elders for recommending appropriate events from the proving information [8]. In addition, it also formed social groups of elders who have similar interests for promoting their social interactions [8]. For elders who like music, an application record favorite songs of elders for gathering them with similar interests, and expanded their social circle by singing songs [9].

Physical devices also used for designing social persuasion functions for elders. Cameras are accessible devices, it provides an easy way for elders as if eating with their families or even unacquainted users, for reducing loneliness feeling and maintaining regular life [10]. A remote presentation system used an easy operation device like a folding monitor, and elders talked with others as if face to face in the real life, including visual and auditory communication [11]. Sharetouch developed a multi-touch social interaction system in a 52-inch touch platform, and provides three types of functions: communication, interactive game and data sharing [12]. Communication function built social network of elders, the game enhanced social interactions among them, and music or photos shared via Bluetooth or internet [12]. A social robot also used to record daily activities of elders, and analyzed their current states from the record of identified behavior, facial expressions, semantics and emotions. In addition, from different interaction behavior of elders, the interaction strategy of elders can be adjusted dynamically for providing appropriate way to accompany them [13]. Hence, an event-triggered system was developed to detect their social behavior in the study.

### 2.2. Danger detection and notification of elders

From researches of alarm systems for elders, systems using in home scenario are discussed in the study. CareBot is an elder care robot to monitor blood pressure and pulse, and a mobile application can send SMS emergency messages with location information to caregivers [14]. Protege is also a mobile application with friendly interface for elders to provide fall detection and SOS notification [15]. Another mobile application also provided monitoring, fall detection and emergency notice using three-axis accelerometer, gyroscope, orientation sensor and GPS in a phone [16]. Moreover, a mobile application provided monitoring function from the collected sensor data of wearable device, and alarm functions differs according to the severity of the situation [17]. In addition, an intelligent surveillance system with learning ability used projected histograms as the main feature of gesture classification for classifying body movements, and sent alarms when detect danger events

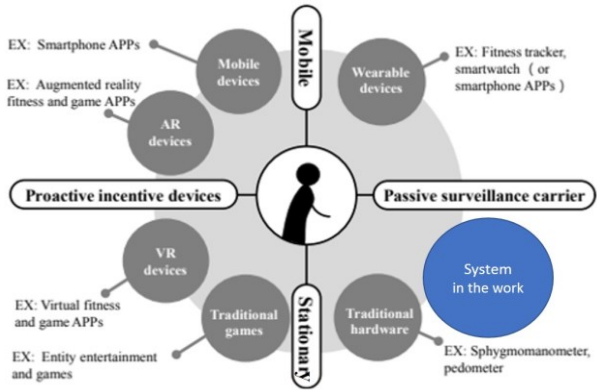

Fig. 2. Types of works in elder care field [26].

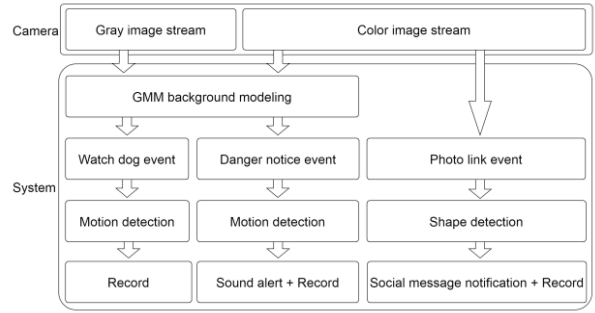

Fig. 3. System architecture.

[18]. Therefore, risks of elders reduce by noticing caregivers in time when accidents happen.

Some researches focused on fall detection or unusual body movements. A study tried to find direct connection between heart disease and fall behavior, and the collected data of heart rate, parallel linear acceleration and position also transmitted to the contact person or hospital [19]. The disadvantage of wearable device is that elders may forget or refuse to bring the device. Using a wall-mounted camera, fall behavior of normal people or elders can be detected from large movements on the MHI image sequence [20]. However, it assumed that there is no movement on the ground after fall behavior and may cause detection error. Another study not only detected fall behavior but also other postures, and feature vectors of approximate ellipsoidal, horizontal and vertical projection histograms around human body and temporal variation of head position were used [21]. The posture information of before and after fall behavior also helps doctors to give diagnosis. In addition, fall speed can identify fall behavior and lying behavior [22].

Environmental sensors or cameras also set for home monitoring. A remote monitoring system combined Arduino Uno, RF transmitter, PC, IP camera and mobile robots, and used the existing application Team Viewer for checking elders and children at home [23]. For privacy concerns of camera setting, system used multiple passive infrared sensors to record routine and detect abnormalities in a long-term monitoring [24]. Although the sensors install easily and have few interruptions to daily lives of people, it cannot recognize multiple users for only judgment from the data of first user. To solve the problem, multiple sensors can be used to record and analyze multiple users, and it requires a complex algorithm and a practical mechanism for the additional cost of data calculation [25]. Hence, an event-triggered system was developed to provide alarm function for avoiding danger situations of elders.

## 3. SYSTEM DESIGN

Current works in elder care field can be classified according to the properties of devices (Figure 2). According to mobility of devices, the devices can be classified into mobile device, for example watches and cellphones and stationary device such as blood glucose machines and cameras. From the data collection way, the categories of devices are proactive incentive devices and passive surveillance carrier. The proactive incentive devices enable elders improving social behavior, and the passive surveillance carrier monitor daily lives of elders passively. In the study, the developed system is classified into stationary and passive surveillance carrier according to the properties of devices, for monitoring danger events of elders in home scenario [26].

### 3.1. System architecture

The system architecture of the study shows in Figure 3. Camera was set in home scenario to detect three types of events: watch dog, danger notice and photo link. Users select 3 ROI regions of the image freely for setting the detection areas of three types of events at home. The freedom also improves accuracy of the system. Once an event is detected, the captured images and time are recorded in text files. Gaussian background modeling was adopted in a gray image sequence for motion detection, and the movements of elders can be identified from the area on the foreground image [27]. When motion detects in the setting area of entrance, triggers the watch dog event. When motion detects in the setting area of danger area such as kitchen or bathroom, triggers the danger notice event and plays a voice alert for reminding elders. From the input of color image, social behavior can be identified using shape recognition. When rectangle shape identifies in the setting area, triggers the photo link event and send social media message to relatives within 5 minutes. Hence, the system has developed for recording both social and danger events of elders.

### 3.2. Detection and record of events

An event-triggered system has developed in the study for recording three types of events: watch dog event, danger notice event and photo link event, representing in blue ROI region, red ROI region and green ROI region, respectively (Figure 1). The system has implemented in C++ program, and calculates the experimental threshold

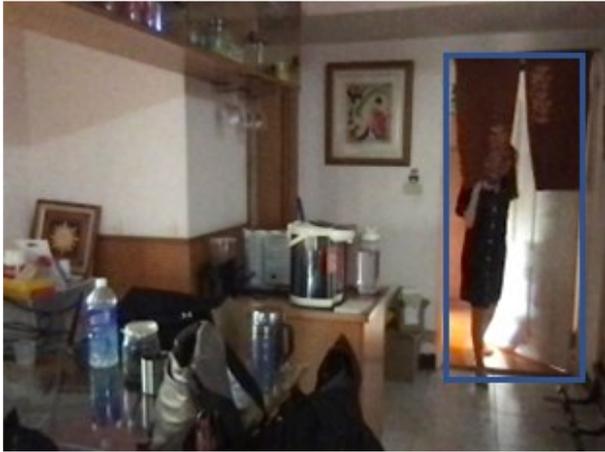

Fig. 4. Watch dog event in home scenario: capture image of family 5.

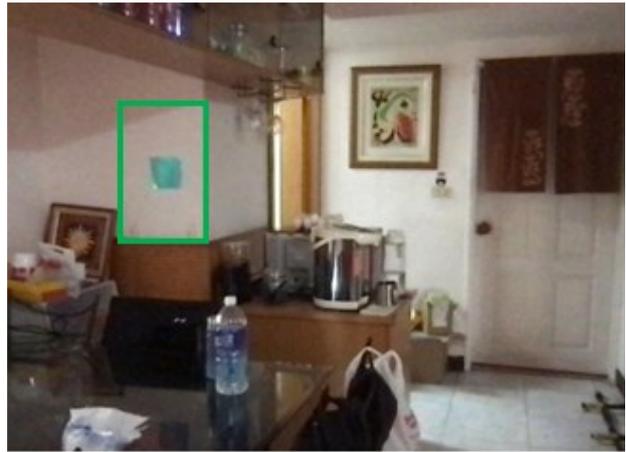

Fig. 6. Photo link event in home scenario: capture image of family 5.

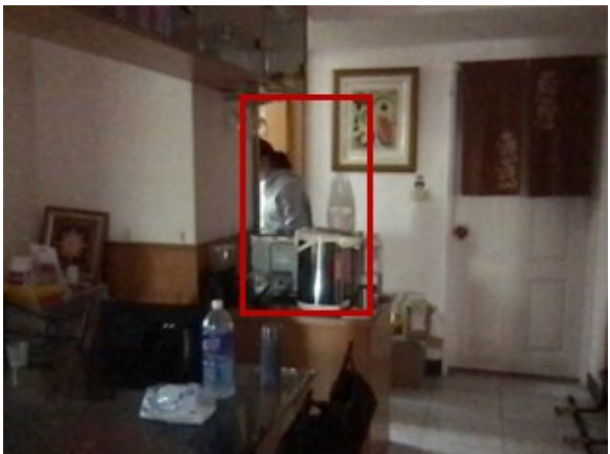

Fig. 5. Danger notice event in home scenario: capture image of family 5.

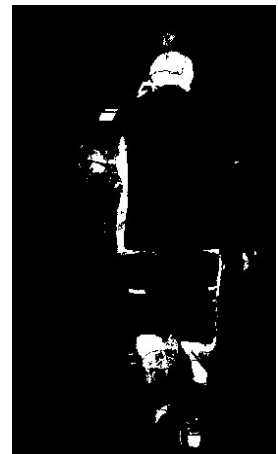
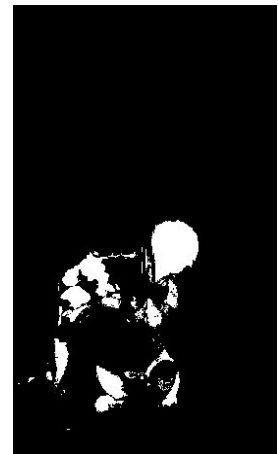

(a)          (b)

Fig. 7. Motion detection using GMM background modeling in watch dog event and danger notice event: (a) normal walking, (b) fall accident.

from the 10 frames in initial stage for accommodating various environment.

### 3.2.1. Watch dog event: real visitors

The detection region of watch dog event often set in the entrance at home. When motion changes over the experimental threshold, the visitors (Figure 4) can be detected and triggers the watch dog event, meanwhile, the captured images and time are recorded (Figure 7 (a)).

### 3.2.2. Danger notice event: entrance of danger area

The detection region of danger notice event often set in the kitchen or bathroom at home. When motion changes over the experimental threshold, the entrance into the danger area can be detected (Figure 5) and triggers the watch dog event, and plays a voice alert for reminding elders to watch the steps. The captured images and time are recorded at the same time (Figure 7 (a)). Both normal behavior and fall accidents are recorder in the system (Figure 7 (a) and (b)).

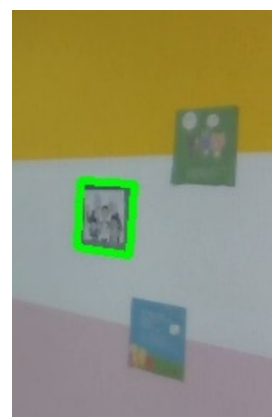
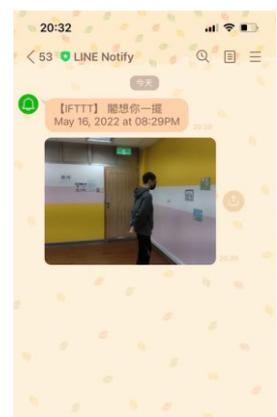

(a)          (b)

Fig. 8. Shape recognition using in photo link event: (a) photo paste behavior, (b) line message notification.

Table 1. Record of three types of events in the preliminary experiment.

| Participants<br>Value | Family 1 | Family 2 | Family 3 | Family 4 | Family 5 |
|---|---|---|---|---|---|
| Experiment days | 4 | 4 | 5 | 4 | 4 |
| Number of watch dog event | 283 | 36918 | 816 | 397 | 333 |
| Number of images in watch dog event | 283 | 36918 | 816 | 397 | 333 |
| Average watch dog event | 70.75 | 9229.5 | 89.8 | 75 | 67.25 |
| Number of danger notice event | 165 | 38862 | 816 | 397 | 333 |
| Number of images in danger notice event | 165 | 38862 | 816 | 397 | 333 |
| Average of danger notice event | 41.25 | 9715.5 | 163.2 | 99.25 | 83.25 |
| Number of photo link event | 2032 | 1974 | 1840 | 1 | 1275 |
| Number of images in photo link event | 2032 | 1974 | 1840 | 1 | 1275 |
| Average of photo link event | 508 | 493.5 | 368 | 0.25 | 318.75 |

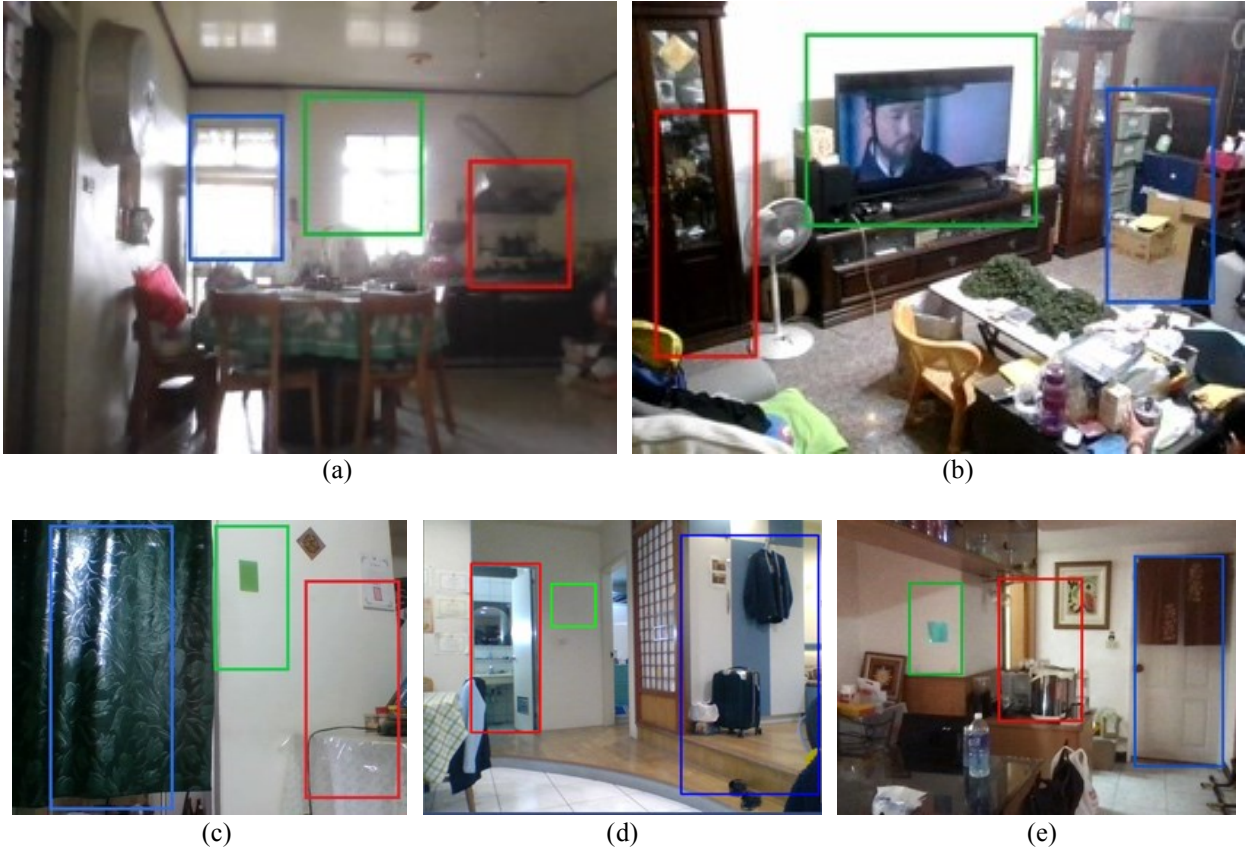

Fig. 9. Real family scenarios in the preliminary experiments:
(a) Family 1, (b) Family 2, (c) Family 3, (d) Family 4 and (e) Family 5

*3.2.3. Photo link event: initiative social behavior*

The detection region of photo link event often set in an aisle on a blank wall at home. An intuitive way was designed for elders to trigger the photo link event by pasting family photos (Figure 6) on the wall as normal life behavior, and then remove the photos on the wall after event. When identifies rectangle shape of an object (Figure 8 (a)), the initiative social behavior from the elders can be detected and triggers the photo link event, and the captured images and time are recorded at the same time. A notification of captured image and a predefined message will be sent to relatives within 5 minutes via social media (LINE notification) (Figure 8 (b)). When relatives or caregivers receive the notification, they are free to contact elders in spare time. Number of active social behavior reflects loneliness degree of elders, and it is an important value for examining psychological condition of elders.

## 4. EXPERIMENT

Experiments in the study contains two parts: preliminary experiment and fall detection experiment. In the preliminary experiments, the system monitored in various environment for testing the robotics of the system in event detection and record. In the fall detection experiments, the collected data was used to train the model for fall detection using SVM (Support Vector Machines) machine learning.

Table 2. Fall pattern and stand pattern after SVM classification

| Participants  Value | Family 1 | Family 2 | Family 3 | Family 4 | Family 5 |
|---|---|---|---|---|---|
| Average prediction of fall pattern | 75.6% | 72.89% | 77.44% | 82.2% | 60.19% |
| Average prediction of stand pattern | 71.43% | 59.89% | 60.78% | 92.42% | 83.41% |
| Average error rate of fall pattern | 77.66% | 31.3% | 81.58% | 74.02% | 76.62% |
| Average error rate of stand pattern | *5.17% | 56.04% | 15.3% | *8.42% | *3.17% |

## 4.1. Preliminary experiments

There are 5 various environments in the preliminary experiments, and the camera setting also differs according to the house space (Figure 6). The embedded camera setting on a laptop covered the detection areas of three types of events, for example a clean wall, or a kitchen or a bathroom. The experiments took at least four hours a day (30, 30, 29, 29 and 34 hours), and 20 participants of 5 families participated in the experiments (age range: 17 to 62 years old; average: 38.25 and SD: 17.13). There is only 1 participant under age 20, 9 participants in the age of 20 to 29; 1 participant in the age of 40 to 49; 7 participants in the age of 50 to 59 years old and 2 participants in the age of 60 to 69.

Table 1 shows the record of three types of events in the 5 various experimental environments, only family 3 took the period of 5 days experiments, and the rest of them all took the period of 4 days experiments. Number of watch dog event and danger notice event reflects the life activities in these families, and pasted photos on purposely to simulate the photo link event for testing the robust of the system. From the experimental results, 2 families (Family 2 and 3) recorded high number of watch dog event, and the number of watch dog event are similar in the rest of families (Family 1, 4 and 5). The circumstances also reflect on the value of danger notice event that 2 families (Family 2 and 3) have high value of danger notice event, and the value of danger notice event are similar in the rest of families (Family 1, 4 and 5). The detection area of family 2 set in the main path of the house (Figure 9 (b)) that cased the obvious high value in the watch dog event, and the reason of high watch dog events in family 3 (Figure 9 (c)) for the camera setting too close to the detection area. To solve the problem of environmental noise, the function of automatic experimental threshold was implanted for improving the system performance. After selection 3 ROI regions of the correlated events, the system calculates the average of experimental threshold from the captured frames within 1 minute. Therefore, the experimental threshold is calculated automatically according to different environments.

In the simulated photo link event, one family (family 4) reflects extremely value that light condition influences the detection results. For example, camera set in a lighting environment (Figure 6 (a)) had better performance than the environment with insufficient light (Figure 6 (d)). 2 families (Family 1 and 2) have high photo link event for the detection area contained rectangle shape of television (Figure 6 (b)) and the rectangle pattern of ceramic tile on the wall (Figure 6 (a)). In addition, number of photo link events larger than the watch dog event and danger notice event. When participants pasted a photo on the wall and did not remove them, the system identified previous rectangle shapes repeatedly. To solve these problems, a method of background subtraction can adopt for improving the system performance in the future work, and only new photos will be recognized without removing old photos.

## 4.2. Fall detection experiments

The same participants joined the fall detection experiments and simulated the fall accidents on purposely. The experiments took 2 hours a day, in the 5 days period. The selection of ROI image at least covered the whole body of participants in the danger notice event. The collected data recorded pattern of standing, walking and falling, and adopted SVM machine learning to train the model. Fall pattern contains the postures of falling and lying on the floor, and stand pattern includes the postures of walking and standing still. After classification data into categories of stand pattern and fall pattern manually, each participants used the data collection of first day as the ground truth to train their models. Each participants classified the rest 4 days data according to the training model of first day data, and classified the collection data into fall pattern category and stand pattern category in different files. Table 2 shows the prediction results of fall pattern and stand pattern after SVM classification. The average of error rate also calculated by checking all the wrong classification manually. From the average prediction value of fall pattern, training results of 4 families (Family 1,2,3 and 4) have good performance. However, the high average error rate of fall pattern represents many images in wrong classification, for falling behavior has various limbs changes than the stand pattern. From the average prediction value of fall pattern, training results of 3 families (Family 1,4 and 5) have good performance, and it also reflects low error rate of stand pattern. From the experimental results, the insufficient training data caused the results in fall pattern classification, and it can be improved by expanding the collection data for model training.

## 5. CONCLUSION

A passive monitoring system for elders using in home scenario has developed in the study. Using technologies of motion detection and shape recognition, three types of events can be detected and recorded. From the experiments of participants from real families, the events recorded from their life activities shows the feasibility of the system. In addition, number of photo link event represents the loneliness state of elders, can be used to calculate loneliness value for examining their mental health in the future work.

## REFERENCES

[1] L. L. Hwang, "Physical, Psychological, and Social Functions of Community-Dwelling Elders Who Are Living Alone," Master's thesis, Kaohsiung Medical University, Taiwan, 2000.

[2] S. W. Tseng, Y. C. Ma and Y.R. Wang, "Social App Design Considerations for Elderly Persons Living Alone Based on User Experience," *Journal of Gerontechnology and Service Management*, Vol. 4, No. 4, pp. 505-520, 2016.

[3] M. C. Shen, "Analysis of the Current Academic Research on the Elderly People Living Alone in Taiwan," Master's thesis, Nan Kai University of Technology, 2019.

[4] Y. C. Wu and S. Y. Chao, "Health Needs, Needs Satisfaction and Correlated Factors among Elderly People Living Alone in Community," *Bulletin on Hungkuang Institute of Technology*, Vol. 63, pp. 44-64, 2011.

[5] 洪健皓、吳聰慧、詹乃嘉、黃政昌 (2020)。「獨居老人心理健康評估」的內涵與方式之探討．諮商與輔導，418，48-53。

[6] M. Z. Syeda and Y. M. Kwon, "Photo Alive! Application and Method for Intergenerational Social Communication," *in Proceedings of the 2017 19th International Conference on Advanced Communication Technology (ICACT)*, 2017.

[7] H. A. Santoso, F. Firdausillah, S. E. Sukmana, A. Yusriana, A. Juliandri and T. Witjahjono, "Berbakti: An Elderly Apps for Strengthen Parent-children Relationship in Indonesia," *in Proceedings of the 2016 International Seminar on Application for Technology of Information and Communication (ISemantic)*, 2016.

[8] J. O. O. Ordoñez, J. F. B. Torres, O. D. S. Villa, E. F. O. Morales, M. L. Nores and Y. B. Fernández, "Stimulating Social Interaction among Elderly People Through Sporadic Social Networks," *in Proceedings of the 2017 International Caribbean Conference on Devices, Circuits and Systems (ICCDCS)*, 2017.

[9] R. Song, Z. Li, X. Yang and Z. Tan, "Music Service System and Intelligent Product Design for the Elderly at Home Based on Social Needs," *in Proceedings of the 2021 2nd International Conference on Intelligent Design (ICID)*, 2021.

[10] D. Korsgaard, T. Bjørner, J. R. B. Pedersen, P. K. Sørensen and F. J. A. P. Cueto, "Eating Together While Being Apart: A Pilot Study on the Effects of Mixed-reality Conversations and Virtual Environments on Older Eaters' Solitary Meal Experience and Food Intake," *in Proceedings of the 2020 IEEE Conference on Virtual Reality and 3D User Interfaces Abstracts and Workshops (VRW)*, 2020.

[11] A. P. Achilleos, C. Mettouris, G. A. Papadopoulos, K. Neureiter, C. Rappold, C. Moser, M. Tscheligi, L. Vajda, A. Tóth, P. Hanák, O. Jimenez and R. Smit, "The Connected Vitality System: Enhancing Social Presence for Older Adults," *in Proceedings of the 12th International Conference on Telecommunications*, 2013.

[12] T. H. Tsai and H. T. Chang, "Sharetouch: A Multi-touch Social Platform for the Elderly," *in Proceedings of the 2009 11th IEEE International Conference on Computer-Aided Design and Computer Graphics*, 2009.

[13] H. Chen, Y. Zhao, Y. Wang, L. Zhao, L. Yin, J. Liu, S. Zhao and G. Chen, "A Framework of Social Robot for Elderly Individuals," *in Proceedings of the 2019 IEEE International Conference on Consumer Electronics - Taiwan (ICCE-TW)*, 2019.

[14] C. Park, S. Kang, J. Kim and J. Oh, "A Study on Service Robot System for Elder Care," *in Proceedings of the 2012 9th International Conference on Ubiquitous Robots and Ambient Intelligence (URAI)*, 2012.

[15] F. Ferreira, F. Dias, J. Braz, R. Santos, R. Nascimento, C. Ferreira and R. Martinho, "Protege: A Mobile Health Application for the Elder-caregiver Monitoring Paradigm," *Procedia Technology*, Vol. 9, pp. 1361-1371, 2013.

[16] G. M. Sung, H. K. Wang and W. T. Su, "Smart Home Care System with Fall Detection Based on the Android Platform," *in Proceedings of the 2020 IEEE International Conference on Systems, Man, and Cybernetics (SMC)*, 2020.

[17] M. Panou, K. Touliou, E. Bekiaris and T. Tsaprounis, "Mobile Phone Application to Support the Elderly," *International Journal of Cyber Society and Education*, Vol. 6, No. 1, pp. 51-56, 2013.

[18] R. Cucchiara, A. Prati and R. Vezzani, "An Intelligent Surveillance System for Dangerous Situation Detection in Home Environments," *Intelligenza Artificiale*, Vol. 1, No. 1, pp. 11-15, 2004.

[19] I. N. Merrouche, A. Makhlouf, N. Saadia and A. R. Cherif, "Overview and Systems of Assistance to Older People," *in Proceedings of the 2016 4th International Conference on Control Engineering & Information Technology (CEIT)*, 2016.

[20] C. Rougier, J. Meunier, A. S. Arnaud and J. Rousseau, "Fall Detection from Human Shape and Motion History Using Video Surveillance," *in Proceedings of the 21st International Conference on Advanced Information Networking and Applications Workshops (AINAW'07)*, 2007.

[21] H. Foroughi, B. S. Aski and H. Pourreza, "Intelligent Video Surveillance for Monitoring Fall Detection of Elderly in Home Environments," *in Proceedings of the 2008 11th International Conference on Computer and Information Technology*, 2008.

[22] A. H. Nasution and S. Emmanuel, "Intelligent Video Surveillance for Monitoring Elderly in Home Environments," *in Proceedings of the 2007 IEEE 9th Workshop on Multimedia Signal Processing*, 2007.

[23] S. Kumar and S. S. Solanki, "Remote Home Surveillance System," *in Proceedings of the 2016 International Conference on Advances in Computing, Communication, & Automation (ICACCA)*, 2016.

[24] Y. Wang, S. Cang and H. Yu, "A Noncontact-sensor Surveillance System Towards Assisting Independent Living for Older People," *in Proceedings of the 2017 23rd International Conference on Automation and Computing (ICAC)*, 2017.

[25] D. Riboni and F. Murru, "Unsupervised Recognition of Multi-Resident Activities in Smart-Homes," IEEE Access, Vol. 8, pp. 201985-201994, 2020.

[26] Y. M. Fang, C. Lin and S. Y. Huang, "The Devices and Interfaces for Elderly Healthcare," *in Proceedings of the 2017 International Conference on Applied System Innovation (ICASI)*, 2017.

[27] C. Stauffer and W. E. L. Grimson, "Adaptive Background Mixture Models for Real-time Tracking," *in Proceedings of the 1999 IEEE Computer Society Conference on Computer Vision and Pattern Recognition (CVPR '99)*, 1999.